# HEURISTICS AND PARSE RANKING[*]

B. Srinivas
Department of Computer Science
University of Pennsylvania
srini@linc.cis.upenn.edu

Christine Doran
Department of Linguistics
University of Pennsylvania
cdoran@linc.cis.upenn.edu

Seth Kulick
Department of Computer Science
University of Pennsylvania
skulick@linc.cis.upenn.edu

**Abstract**

There are currently two philosophies for building grammars and parsers – Statistically induced grammars and Wide-coverage grammars. One way to combine the strengths of both approaches is to have a wide-coverage grammar with a heuristic component which is domain independent but whose contribution is tuned to particular domains. In this paper, we discuss a three-stage approach to disambiguation in the context of a lexicalized grammar, using a variety of domain independent heuristic techniques. We present a training algorithm which uses hand-bracketed treebank parses to set the weights of these heuristics. We compare the performance of our grammar against the performance of the IBM statistical grammar, using both untrained and trained weights for the heuristics.

## 1 Introduction

Although statistical POS taggers [Church1988] have conclusively out performed hand-crafted POS taggers [Harris1962; Klein and Simmons1963] of the pre-statistical NLP era, the same cannot be said about parsers and the grammars they use. There are currently two philosophies for building grammars and parsers. Wide-coverage grammars (WCGs) such as [Grover et al.1993; Alshawi et al.1992; Karlsson et al.1994; Group1995] are mostly hand-crafted and are designed to be domain-independent. They are usually based on a particular grammar formalism, such as CFG, Dependency Grammar or LTAG. Statistically induced grammars (SIGs) [Jelinek et al.1994; Magerman1995; Mori and Nagao1995] on the other hand, are trained on specific domains using a manually annotated corpus of parsed sentences from the given domain.

Aside from the methodological differences in grammar construction, the two approaches differ in the richness of information contained in the output parse structure. Wide-coverage grammars generally provide a far more detailed parse than that output by a statistically induced grammar. Also, the linguistic knowledge which is overt in the rules of hand-crafted grammars is hidden in the statistics derived by probabilistic methods, which means that generalizations are also hidden and the full training process must be repeated for each domain.

Ideally, one would like to combine the strengths of both approaches, by having a rule-based grammar with a heuristic component which is domain independent but whose contribution is tuned to particular domains. One way to capture the strengths of SIGS of associating preferences with parses in a WCG, is to introduce heuristics to rank parses. Both domain independent and domain dependent heuristics can be included in WCGs. In contrast, a SIG does not distinguish between grammatical knowledge, knowledge of domain independent and domain dependent preference.

Researchers [Hobbs and Bear1994; McCord1993] have suggested domain-independent heuristics in the context of CFGs. However, in recent grammar formalisms the lexicon has come to play a more central role than it does in non-lexicalized CFGs. These grammars are 'lexicalized' in that each elementary structure of the grammar is associated with at least one lexical item. These structures provide a domain over which syntactic and semantic constraints can be specified by the lexical item that selects them. It is precisely the power of these "extended domains of locality" which makes lexicalized grammars so well-suited to the specification of natural language grammars. In this paper, we show that the lexicalized grammars with extended domains of locality provide a unique opportunity to apply known disambiguation techniques in an efficient manner, and in particular to exploit lexically sensitive heuristics that cannot be stated easily on a non-lexicalized grammar.

[*]The authors would like to thank Aravind Joshi, Anoop Sarkar, Joseph Rosenzweig, Ted Briscoe and Henry Thompson for their valuable comments and assistance with this work. The work was partially supported by ARO grant DAAL03-89-0031, ARPA grant N00014-90-J-1863, NSF STC grant DIR-8920230, and Ben Franklin Partnership Program (PA) grant 93S.3078C-6.

In this paper we discuss a three-stage approach to disambiguation in parsing, with particular emphasis on lexicalized grammars. The first stage is part-of-speech tagging on the input sentence. By ascertaining the part-of-speech of each word, one is able to greatly reduce the number of structures (trees) to be considered in the parsing process itself. The second opportunity for disambiguation comes when a set of possible structures has been selected for each lexical item. At this point certain structures can be filtered out, using statistical methods and heuristics which take advantage of the shapes of the structures. Finally, any remaining ambiguous parses are ranked using further heuristics.

We then discuss a method for combining the heuristics and automatically training their weights. We use hand-bracketed treebank parses for the training process. These trained heuristics are then used to evaluate the performance of our system against the IBM probabilistic grammar (on IBM-manual data). While it may appear that disambiguation and evaluation are disparate issues, we believe that it is useful to consider them together insofar as the results of the disambiguation step are the input to any evaluative processes.

The particular system we are using for our experiments is a wide-coverage, domain-independent system based on lexicalized Tree Adjoining Grammar (LTAG). We will assume that readers are familiar with the formalism. See [Joshi et al.1975], [Vijay-Shanker1987], [Vijay-Shanker and Joshi1991] and [Schabes et al.1988] for description of TAG and lexicalized TAG. A summary of each component of the XTAG system [Group1995] is presented in Table 1.

## 2  Disambiguation Techniques

There are a number of stages where syntactic ambiguity – lexical and structural – can be reduced in parsing. We will discuss: (1) part-of-speech tagging prior to parsing, (2) tree/subcat[1] filtering and weighting techniques, and (3) heuristics to rank the generated parses. The first and last are general techniques which are applicable to all types of grammars; tree filtering and tree weighting take advantage of the particular properties of lexicalized grammars. The combination of these three techniques has proven extremely effective in attacking the problem of ambiguity while simultaneously improving the efficiency of the parser.

### 2.1  Part-of-speech Tagging

It is well known that lexical ambiguity with regard to part-of-speech (POS) is one of the greatest sources of overall ambiguity. This is particularly important in a lexicalized grammar, where each word is associated with multiple structures for each POS it may have. In our grammar, for example, the word *try* selects 59 verb trees and 17 noun trees; simply by identifying its POS we substantially reduce the number of trees it contributes to the parsing process. When this is done for each word in a sentence, the reduction in number of trees selected is enormous. Consider two examples: the NP *the act of allowing fresh air into a room* receives 26 parses untagged, and POS tagging reduces that number to 4; the sentence *the second part is the name of your personal computer* receives 32 parses without POS tagging, and only 8 parses with tagging.[2]

### 2.2  Tree Filtering and Tree Weighting Techniques

**Filtering**  The second opportunity for disambiguation comes after trees have been selected for each word of the input sentence, in the first pass of the parser. Structural properties of trees such as the span of the tree and the position of the anchor in the tree are used to weed out unsuitable trees. This has the effect of eliminating, for example, a noun tree with a determiner position when there is no determiner to the left of the noun in the input string. At this stage we have eliminated trees which would never be used in any parse of the input string, thus reducing the initial search space as early as possible for the parser. This does not however reduce the number of parses produced.

---

[1] Henceforth, we will just say "trees" but the reader should bear in mind that these techniques are equally applicable to other lexicalized grammars, whatever the structure they associate with each lexical item.

[2] The first example is from the Alvey test sentences and the second from the IBM Manual Corpus.

| Component | Details |
| --- | --- |
| Morphological Analyzer and Morph Database | Consists of approximately 317,000 inflected items. Thirteen parts of speech are differentiated. Entries are indexed on the inflected form and return the root form, POS, and inflectional information. Database does not address derivational morphology. |
| POS Tagger and Lex Prob Database | Wall Street Journal-trained trigram tagger [Church1988] Decreases the time to parse a sentence by an average of 93%. |
| POS Blender | Combines information from the Morphology and the POS tagger Outputs N-best POS sequences with morphological information for each word of the input sentence. |
| Tree Database | 566 trees, divided into 40 tree families and 62 individual trees. Tree families represent subcategorization frames. E.g., the intransitive tree family contains the following trees: indicative, wh-question, relative clause, imperative and gerund. Individual trees are generally anchored ($\diamond$) by non-verbal lexical items that substitute or adjoin into the clausal trees. Feature values may be specified within a tree or may be derived from the syntactic database. |
| Syntactic Database and Statistical Database | Associates lexical items with the appropriate trees and tree families based on subcategorization information. Extracted from OALD and ODCIE and contains more than 105,000 entries. Each entry consists of: the uninflected form of the word, its POS, the list of trees or tree-families associated with the word, and a list of feature equations that capture lexical idiosyncrasies. |
| X-Interface | Menu-based facility for creating and modifying tree files. User controlled parser parameters: parser's start category, enable/disable/retry on failure for POS tagger. Storage/retrieval facilities for elementary and parsed trees as text and postscript files. Graphical displays of tree and feature data structures. Hand combination of trees by adjunction or substitution for diagnosing grammar problems. |

Table 1: **XTAG System Summary**

Furthermore, statistical information about the usage frequency of trees has been collected by parsing corpora. This information has been compiled into a database that is used by the parser. The database contains tree unigram frequencies of trees collected by parsing the Wall Street Journal, IBM manual, and ATIS corpora. The top 15 trees for each of these corpora are shown in Table 2. It is interesting to note that the *PP_Attaches_to_NP* structure ranks second in ATIS and ranks lower in IBM Manual and WSJ corpora. The parser, augmented with the statistical database, assigns each word of the input sentence the three most frequently used trees for that word. Statistical filtering removes trees which might ultimately participate in a less likely parse. Note that this is different from inducing the grammar from a particular corpus, in that the database contains a composite of frequencies. On failure the parser retries using all the trees suggested by the syntactic database for each word.[3] The augmented parser succeeds in parsing 50% of input sentences using only the top three trees for each word.

| #  | ATIS                 | Prob.   | IBM Manual           | Prob.   | WSJ                 | Prob.   |
|----|----------------------|---------|----------------------|---------|---------------------|---------|
| 1  | Noun_Phrase          | (0.260) | Determiner           | (0.175) | Noun_Phrase         | (0.184) |
| 2  | PP_Attaches_to_NP    | (0.142) | Noun_with_Det        | (0.174) | Noun_Mods_Noun      | (0.151) |
| 3  | Noun_Mods_Noun       | (0.105) | Noun_Mods_Noun       | (0.112) | Determiner          | (0.089) |
| 4  | Noun_with_Det        | (0.094) | Aux_Verb             | (0.095) | Noun_with_Det       | (0.079) |
| 5  | Determiner           | (0.092) | Noun_Phrase          | (0.073) | Aux_Verb            | (0.074) |
| 6  | Imperative_Double_Obj| (0.056) | Adjective            | (0.044) | Adjective           | (0.055) |
| 7  | Aux_Verb             | (0.050) | PP_Attaches_to_VP    | (0.041) | PP_Attaches_to_VP   | (0.038) |
| 8  | Adjective            | (0.045) | Passive_Trans        | (0.037) | PP_Attaches_to_NP   | (0.027) |
| 9  | Inverted_Aux         | (0.044) | Indic_Transitive     | (0.035) | PRO                 | (0.025) |
| 10 | Extracted_Predicative| (0.030) | PP_Attaches_to_NP    | (0.033) | Pre-VP_Adverb       | (0.021) |
| 11 | Imperative_Transitive| (0.021) | Imperative_Transitive| (0.015) | Indic_Transitive    | (0.021) |
| 12 | Stacked_Det          | (0.011) | PRO                  | (0.012) | Indic_Scomp         | (0.014) |
| 13 | Obj_Extr_Trans       | (0.008) | VP_Negation          | (0.009) | Rel_Cl_Transitive   | (0.011) |
| 14 | PP_Attaches_to_VP    | (0.007) | Indic_Intrans        | (0.008) | Sent_Adv            | (0.010) |
| 15 | Comp_Extr_SComp      | (0.004) | Post-VP_Adverb       | (0.007) | Post-VP_Adv         | (0.010) |

Table 2: **The Unigram probabilities of top 15 trees from ATIS, IBM Manual and WSJ Corpora**

Continuing with an example discussed above, *the second part is the name of your personal computer*, the number of parses is reduced from 8 to 3 by the tree-filtering techniques. In addition, these methods speed the overall runtime by a factor of 8.

**Local Weighting** Having selected the top three trees for each word in the sentence, we add general (dis)preference weightings for clause types. Some sample preference heuristics are:

1. Disprefer relative clause trees.
2. Disprefer topicalization.
3. Disprefer predicative trees.

We also include certain lexical preferences for highly ambiguous function words, such as:

1. Prefer *of* as NP over VP modifier.
2. Prefer *this* as Determiner over Noun.
3. Prefer *to* as Verb over Preposition.
4. Prefer *that* as Complementizer over Determiner.
5. Prefer *which* as Complementizer over Noun.

The four parses remaining for our example after tree filtering are now passed onto the next stage of processing with their associated weights.

---

[3]We are currently working on an agenda-based parser, which would prefer higher ranked trees and would not require such a fall-back strategy.

## 2.3 Global Heuristics for Ranking the Generated Parses

Any parses which survive POS tagging and tree-filtering are generated by the second stage of the parser in ranked order. This ranking is determined using heuristics based on structural preferences in the derivation. These preferences include:

1. Prefer argument positions to adjunct positions (here, this amounts to preferring fewer adjunction operations).

2. Prefer low-attached PPs.

3. Prefer high-attached Adjectives.

These heuristics differ in scope from the more local preferences described in the previous section, which are applied in the first stage of the parser.

For our example sentence, the heuristics rank highest the same sentence which human judges also selected as the correct parse (parse #3). The rule governing PP attachment is clearly the most important for this sentence. The three ranked sentences are shown schematically in Figure 1. The ranking of parses facilitates selection of a number of parses to be passed on to further levels of processing. For purposes of evaluation, we can select the number of parses equal to the number considered in the system being evaluated against. In applications emphasizing speed, only the highest-ranked parse will be considered. In applications emphasizing coverage, the top $n$ parses can be considered.

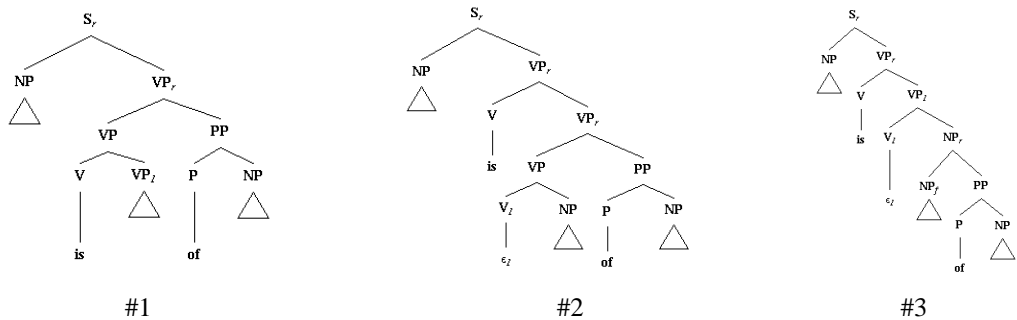

Figure 1: **Ranked Parses Generated For Sample Sentence**
*The second part is the name of your personal computer*

## 2.4 Discussion of Heuristics

Similar heuristics have been used by (non-lexicalized) CFG-based parsers (e.g. [Hobbs and Bear1994], [McCord1993], and [Nagao1994]). In this section we highlight the differences between lexicalized grammars and CFGs in terms of the methodology and applicability of the heuristics.

Local heuristics which refer to single lexical items or entire clausal constructions are easily stated within a lexicalized grammar, where they refer to a single constituent. Also lexical sensitivity can be exploited quite elegantly in the rules of a lexicalized grammar where the entire argument structure of a given lexical item is localized within an single rule. In a CFG, for example, the argument structure would be dispersed across a number of rules and would be more difficult to identify.

The weights of both sets of heuristics are combined by a linear function that is at first hand-set and then altered by the heuristics training described below. As we will discuss in the experimental section, we have developed a means of automatically setting these weights using a training algorithm. This technique give us the advantages of statistical methods in identifying structural generalizations in texts, while retaining the linguistically transparency and ease of portability of a rule-based system.

The heuristics themselves are entirely domain independent; only the weights are affected by training. When working with a particular domain, one could easily augment these general heuristics with more specialized ones. In this way, the heuristics have an advantage over purely statistical approaches which must be completely retrained for each new genre. Hobbs and Bear [1994] have proposed two general heuristics but do not suggest of a way to combine or evaluate them in any detail. McCord's [1993] heuristics are domain independent but the contributions of the heuristics are hand-tuned for the computer-manual domain.

Alshawi and Carter [1994], working with a non-lexicalized CFG-based large scale grammar, specialize their grammar to the ATIS domain using heuristics whose contributions are trained on hand-picked correct Quasi-Logical Forms. They find that heuristics based on lexical semantic collocations dominate other heuristics in selecting the correct parse. However, these heuristics are domain dependent and need to be computed for each domain. Based on their results, we expect that adding semantic collocational information to our heuristic set would improve our results. Like our current heuristics, we will only add collocations which are stable across several genre of text.

## 3 Evaluation

As discussed above, the ranking of parses for each sentence is crucial to evaluating the grammar against other systems. XTAG has recently been used to parse Wall Street Journal[4], IBM manual, and ATIS corpora as a means of evaluating the coverage and correctness of XTAG parses.

### 3.1 Coverage

To evaluate the coverage of our grammar, a sentence is considered to have parsed if XTAG produces any parses. Verifying the presence of the correct parse among the generated parses is done manually at present by random sampling. Results without the use of parse ranking are shown in Table 3.[5] It is worth emphasizing that the XTAG grammar is truly wide-coverage and has not been fine-tuned to any particular genre, unlike many other grammars.

| Corpus | # of Sentences | % Parsed | Av. # of Parses/Sent |
|---|---|---|---|
| WSJ | 18,730 | 41.22 % | 7.46 |
| IBM Manual | 2040 | 75.42% | 6.12 |
| ATIS | 524 | 88.35% | 6.0 |

Table 3: **Performance of XTAG on various corpora**

Performance on the WSJ corpus is lower relative to IBM and ATIS due to the wide-variety of syntactic constructions present. Even grammars induced on the partially bracketed WSJ corpus have fairly low performance (e.g. 57.1% sentence accuracy for [Schabes et al.1993]).

### 3.2 Correctness

The second aspect of evaluating our grammar is determining whether we obtain the correct parse, as defined by a hand-bracketed corpus. We use a crossing bracket measure for evaluation. Crossing brackets is the percentage of sentences with no pairs of brackets crossing the treebank bracketing (i.e. ( ( a b ) c ) has a crossing bracket measure of one if compared to ( a ( b c ) ) ). Recall is the ratio of the number of constituents in the XTAG parse to the number of constituents in the corresponding Treebank sentence. Precision is the ratio of the number of correct constituents to the total number of constituents in the XTAG parse.

**Experiment 1: Equally Weighted Heuristics** A more detailed experiment to measure the crossing bracket accuracy of the XTAG-parsed IBM-manual sentences has been performed; performance results from other parsers on WSJ or ATIS are not available. In this experiment, XTAG-parses of 1100 IBM-manual sentences have been ranked using the heuristics discussed in this paper, with all heuristics weighted equally. The ranked parses have been compared[6] against the bracketing given in the Lancaster Treebank of IBM-manual sentences[7]. Table 4 shows the results of XTAG obtained in this experiment, which used the highest ranked parse for each system. It also shows the results of the latest IBM statistical grammar [Jelinek et al.1994] on the same genre of sentences. Only the highest-ranked parse of both systems was used for this evaluation.

---
[4]Sentences of length $\leq$ 15 words.
[5]This result was previously reported in [Doran et al.1994].

| System | # of sentences | Zero Crossing Bracket % | Recall % | Precision % |
|---|---|---|---|---|
| XTAG | 1100 | 81.29% | 82.34% | 55.37% |
| IBM Statistical grammar | 1100 | 86.20% | 86.00% | 85.00% |

Table 4: **Performance of XTAG on IBM-manual sentences**

As can be seen from Table 4, the precision figure for the XTAG system is considerably lower than that for IBM. For the purposes of comparative evaluation against other systems, we had to use the same crossing-brackets metric though we believe that the crossing-brackets measure is inadequate for evaluating a grammar like XTAG. There are two reasons for the inadequacy. First, the parse generated by XTAG is much richer in its representation of the internal structure of certain phrases than those present in manually created treebanks (e.g. IBM: [$_N$ your personal computer], XTAG: [$_{NP}$ [$_G$ your] [$_N$ [$_N$ personal] [$_N$ computer]]]). The detailed bracketing provided by XTAG make the parse structure more informative. This is reflected in the number of constituents per sentence, shown in the last column of Table 5. We are aware of the fact that increasing the number of constituents also increases the recall percentage.

We measured the number of constituents per parse with the internal structure of NPs and VPs removed to more closely correspond to the Treebank parse. Precision improved to 66.64% while recall dropped slightly to 80.20%. We believe that the precision percentage can be further improved by flattening all adjunction structures (such as PPs, Adverbs and other modifiers) since each adjunction in LTAG adds an extra level of structure.

| System | Sent. Length | # of sent | Av. # of words/sent | Av. # of Constituents/sent |
|---|---|---|---|---|
| XTAG | 1-10 | 654 | 7.45 | 22.03 |
| | 1-15 | 978 | 9.13 | 30.56 |
| IBM Stat. Grammar | 1-10 | 447 | 7.50 | 4.60 |
| | 1-15 | 883 | 10.30 | 6.40 |

Table 5: **Constituents in XTAG parse and IBM parse**

A second reason for considering the crossing bracket measure inadequate for evaluating XTAG is that the primary structure in XTAG is the derivation tree. Two identical bracketings for a sentence can have completely different derivation trees (e.g. *kick the bucket* as an idiom vs. a compositional use). A more direct measure of the performance of XTAG would evaluate the derivation structure, which captures the dependencies between words.

**Experiment 2: Weighted Heuristics** For Experiment 2, an iterative process is used to train the heuristics. This is similar in spirit to the the work presented in [Alshawi and Carter1994], although the details of the training algorithm differ. In addition, the training algorithm in [Alshawi and Carter1994] used semantic representation for sentences restricted from ATIS domain as the "gold" corpus to train on. We experiment with phrase-structure parses of sentences from IBM-manual data and our training algorithm uses hand-bracketed Lancaster Treebank parses as the "gold" corpus.

---
[6]We used the Parseval program written by Phil Harison (phil@atc.boeing.com).
[7]The Treebank was obtained through Salim Roukos (roukos@watson.ibm.com) at IBM.

An IBM-manual corpus of 931 sentences was randomly split into three groups: TRAIN - 626 sentences, HELD-OUT - 205 sentences, and TEST - 100 sentences. TRAIN is used to train new heuristic weights, while HELD-OUT is used as a control to prevent overtraining on TRAIN. TEST is set aside to be used as a test for the final weightings. We use the same gold standard as in experiment 1. Each iteration chooses, at random, one of the heuristics to adjust, and a random amount to adjust it by. All of the sentences in TRAIN are then ranked using the adjusted heuristic (with the other heuristics unchanged).

The top six parses by this ranking are then compared with the parse from the gold standard. The crossing bracket, recall, and precision measures computed by Parseval are all taken equally into account when determining whether a heuristic change resulted in an improvement. If the result of this comparison (for all the sentences in TRAIN as a whole) is an improvement, then the random change is kept. Sentence group HELD-OUT is used to determine when this iterative process ends. Each time a heuristic change results in an improvement, all of the sentences in HELD-OUT are ranked using the new heuristic settings, and the overall result is compared to the last score for HELD-OUT. The iteration terminates when there has been no improvement three consecutive times.

Table 6 shows the results of this training process. The sentence group column indicates the set of sentences (HELD-OUT/TEST) on which the performance was measured. The Experiment column indicates the three sets of experiments performed on each of the sentence sets. The performance in all the experiments has been measured using the crossing bracket metric on the top six parses. The metric crossing bracket accuracy measures the percentage number of sentences with no crossed brackets. Crossing bracket average gives the average number of crossed bracket errors per parse.

| Sentence Group | Experiment | Zero Crossing Bracket % | Crossing Bracket Average | Recall % | Precision % |
|---|---|---|---|---|---|
| HELD-OUT | No heuristics | 77.56 | 1.15 | 81.57 | 54.01 |
| | No preference | 81.46 | 1.08 | 82.42 | 54.50 |
| | Preferences Trained I | 83.41 | 1.05 | 83.46 | 55.17 |
| | Preferences Trained II | 83.90 | 1.03 | 83.71 | 55.33 |
| TEST | No heuristics | 85.00 | 1.03 | 80.87 | 54.45 |
| | No preference | 87.00 | 0.97 | 82.11 | 55.26 |
| | Preferences Trained I | 88.00 | 0.96 | 82.61 | 55.59 |
| | Preferences Trained II | 89.00 | 0.94 | 82.98 | 55.83 |

Table 6: **Performance Results on the Crossing Bracket Measure for the HELD-OUT and TEST set without any heuristics, with heuristics but with no preference weights, and with heuristics weighted by weights set by the training process.**

1. No heuristics: This experiment refers to the baseline performance of the system in which all the parses were ranked equally and in case there are more than six parses, the first six on the list of parses were chosen for measuring the performance. The results are shown in the first and the fourth rows for the HELD-OUT and the TEST set respectively.

2. No preference: In this experiment, parses were ranked using the heuristics, however, all the heuristics had equal contributions to the rank of a parse. The results are shown in the second and the fifth rows for the HELD-OUT and the TEST set respectively. Including heuristics has improved the performance when compared to the performance with no heuristics.

3. Preferences Trained: In this experiment performance was measured on the top six parses that were ranked using the heuristics weighted according to the weights set by the training process. In Table 6 we show the results of the performance using the weights resulting from two training runs.

### 3.3  Discussion of the Experiments

As can be seen from the "Preferences Trained II" line in the "Test" data from Table 6, using our automatic training method we outperform the IBM statistical grammar result, given in Table 4. The training algorithm improves the performance of the heuristics slightly. We believe that this is due to the domain-independent nature of the heuristics which makes the training algorithm less crucial than it might be if the heuristics included were domain-dependent. We are currently investigating this issue in the context of WSJ data.

We are collecting a hand-checked corpus of XTAG parsed sentences that includes the derivation trees which we will be able to use as a "gold" standard for future experiments. This will enable us to replace the crossing bracket measure with an exact match metric for the training procedure.

These experiments show that with domain independent heuristics the performance of a wide-coverage grammar can equal or better the performance of a statistically induced grammar. There is further opportunity for improvement by adding domain-dependent heuristics, in particular semantic collocations. However, in the interest of portability, we will not include such information in the present system.

## 4 Future Work

The heuristics presented in this paper do not completely exploit the lexical sensitivity provided by the LTAG representation. Towards incorporating more lexical information, we intend to use lexical collocation information as one of the factors for disambiguation. We are in the process of collecting lexical collocation information that is reasonably domain independent from various corpora. We expect that including the such information will improve the results significantly.

We will continue to add and test the effectiveness of new heuristics, and to continue to refine our training algorithm. In addition, we plan to evaluate our grammar using the same techniques on other genre of text, in particular Wall Street Journal data.

As mentioned earlier, we feel that crossing bracket metric is not an adequate for measuring the performance of the grammar formalisms such as LTAG. Hence it serves as a poor objective function to improve the performance of such grammars. We are forced to using the crossing bracket measure since the only corpora available at present that serve as gold standard are ones that are constituent-bracketed. To overcome this problem, we are collecting an LTAG-parsed corpus with each sentence annotated with its correct derivation tree. This resource will prove invaluable for future research in training disambiguation heuristics.

## 5 Conclusion

In this paper, we have discussed a three-stage approach to disambiguation in the context of a lexicalized grammar, using a variety of domain independent statistical techniques. We have presented a training algorithm which uses hand-bracketed treebank parses to set the weights of these heuristics. We show that the performance of our grammar is comparable to the performance of the IBM statistical grammar, using both untrained and trained weights for heuristics.